\documentclass[11pt]{article}

\usepackage[final]{acl}

\usepackage{times}
\usepackage{latexsym}

\usepackage[T1]{fontenc}

\usepackage[utf8]{inputenc}

\usepackage{microtype}

\usepackage{inconsolata}

\usepackage{graphicx}

\title{Enhancing LLMs in Predictive Political QA with Semi-Structured Data}

\author{
\textbf{Yinan Liu}\textsuperscript{*,\textdagger} \quad
\textbf{Zihan Zhou}\textsuperscript{\textdagger} \\
\textbf{Zichun Jin} \quad
\textbf{Xinyu Wang} \quad
\textbf{Bin Wang} \quad
\textbf{Xiaochun Yang} \\
School of Computer Science and Engineering, Northeastern University, Shenyang 110819, China \\
\textsuperscript{\textdagger}Equal contribution. \textsuperscript{*}Corresponding author.}

\usepackage{algorithm}
\usepackage{xcolor} 

\definecolor{zzhpurple}{RGB}{128,0,128}

\usepackage{amsmath}
\usepackage{multirow}
\usepackage{booktabs}

\usepackage{pgfplots}
\pgfplotsset{compat=1.17} 
\usepackage{pgfplotstable}
\usepackage{tikz}
\usepackage{subcaption}
\usetikzlibrary{shapes,positioning}
\usepackage{pgfkeys}  
\usepackage[skins,breakable]{tcolorbox}  
\usepackage{lipsum} 
\usepackage{graphicx}

\newcommand{\promptbox}[2]{%
  \par\noindent
  \begin{minipage}{\linewidth}
    \centering
    \begin{tikzpicture}
      \node[
        draw=brown!80!black,
        rectangle split,
        rectangle split parts=2,
        rectangle split part fill={brown!60,orange!20},
        rectangle split part align={center,left},
        rounded corners=3pt,
        text width=1\linewidth,
        inner sep=4pt,
        font=\small
      ] {
        \textbf{#1}
        \nodepart{two}
        #2
      };
    \end{tikzpicture}
  \end{minipage}%
  \par
}

\definecolor{lightblue}{RGB}{240, 248, 255} 
\definecolor{darkblue}{RGB}{55, 113, 190}

\newtcolorbox{takeaway}[1][]{%
  enhanced,
  breakable,
  colback=lightblue,
  colframe=lightblue,
  width=\linewidth,
  left=4mm,
  right=3mm,
  top=2mm,
  bottom=2mm,
  boxrule=0pt,
  borderline west={3pt}{0pt}{darkblue},
  #1
}

\begin{document}

\maketitle

\begin{abstract}

Predictive political question answering (QA), such as predicting how a political actor will vote, goes beyond factual lookup. 
External political resources offer rich historical evidence, but rarely contain the answer itself.
Existing LLM augmentation methods, including actor-profile-based simulation and knowledge graph evidence injection, improve political reasoning but largely treat external resources as knowledge-based evidence, leaving prediction-relevant signals under-modeled.
We identify two complementary signals for predictive political QA: actor stances that capture issue-specific preferences, and high-order structure signals that capture indirect dependencies among political actors.
We propose PSL, a dual-view framework that converts semi-structured political records into inference-oriented evidence for LLMs. 
PSL extracts stance signals from question-relevant actor records in a semantic view, and learns structure-aware actor representations from an actor interaction graph in a vector view. 
Across three real-world datasets and multiple LLMs, PSL consistently outperforms baselines, with ablations confirming the complementary gains of stance and structure signals.

\end{abstract}

\section{Introduction}

Political question answering (QA) serves diverse information needs about political actors, policies, and events. 
Large language models (LLMs) provide a powerful foundation for political QA, but their knowledge is often incomplete and unreliable in specialized, long-tail settings~\cite{hallucination}.
Augmenting LLMs with external political resources is therefore a natural direction~\cite{rag1}.
The political domain contains abundant external data, including social background information\footnote{https://www.wikipedia.org/}, electoral records\footnote{https://ballotpedia.org/}, and legislative voting records\footnote{https://legiscan.com/}, much of which is semi-structured or structured. 
However, many political QA tasks are not simple factual queries but predictive questions, such as predicting how a political actor will vote or respond to a future event~\cite{yang2021joint, galla2018predicting}.
For these questions, external resources provide valuable records but rarely contain the answer directly, because the relevant behavior has not yet occurred and existing records serve only as indirect facts. 
Therefore, predictive political QA cannot be solved by direct factual retrieval alone.

\begin{figure*}[t]
\centering
\includegraphics[width=0.88\textwidth]{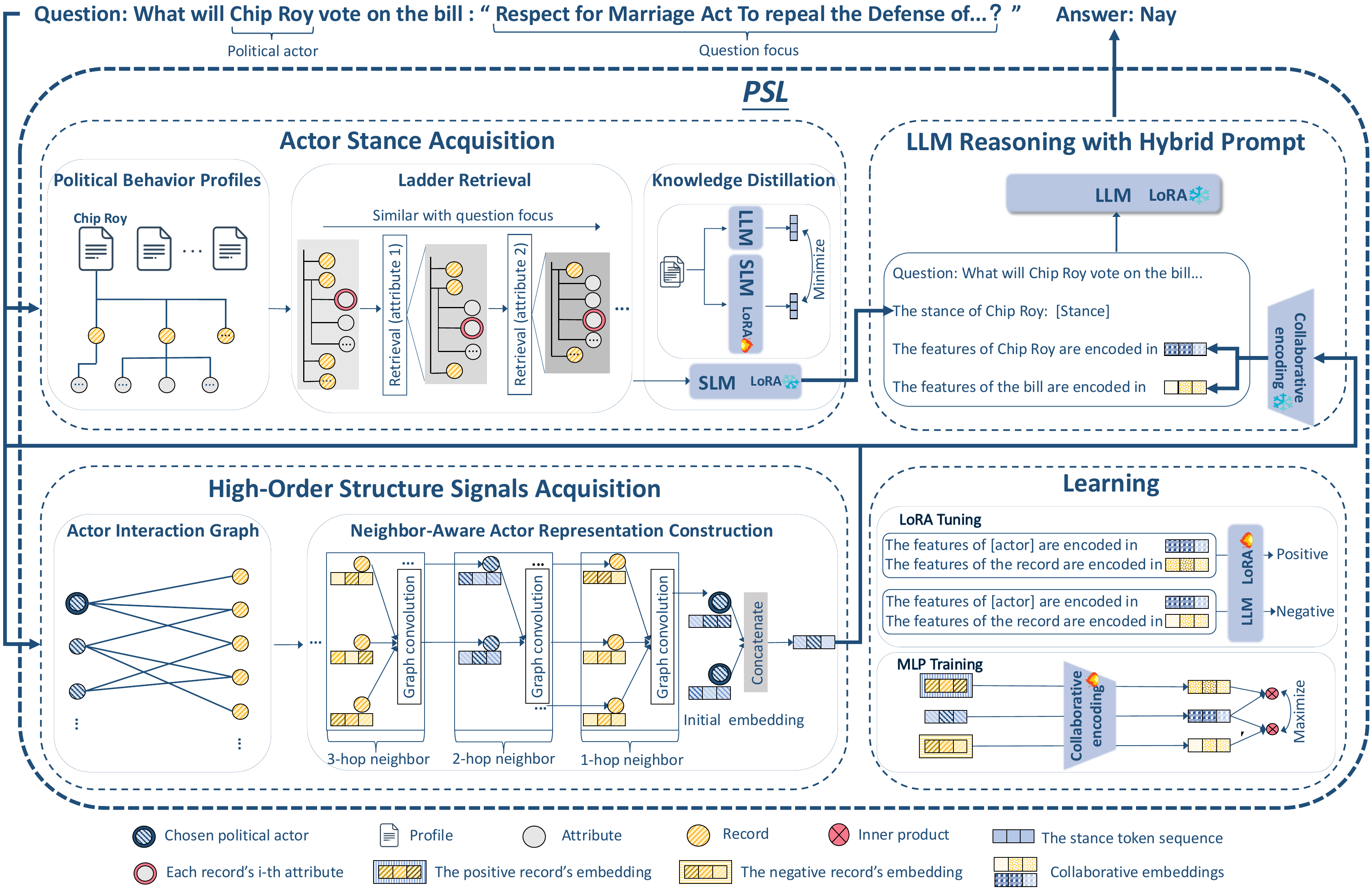} 
\caption{Overview of our framework PSL.}
\label{mainfig2}
\end{figure*}

To the best of our knowledge, the closest recent studies are PAA~\cite{li2025political} and PEG~\cite{mou2024unifying}, which enhance LLMs’ political reasoning with external resources.
PAA uses actor-profile-based simulation, organizing actors’ backgrounds and past behavior into textual profiles and prompting LLMs to simulate their decisions through role-playing. 
PEG uses knowledge-graph-based evidence injection, converting external resources into a knowledge graph (KG) and retrieving relevant structured evidence to augment LLMs.
Both show the value of external resources for political reasoning, but still largely treat them as knowledge-based evidence (\textit{e.g.} Political Behavior Profiles in Fig.~\ref{mainfig2}). 
Moreover, profile-based methods struggle to capture implicit group-level influence among political actors~\cite{sowden2018quantifying}, whereas KG-based methods often lose fine-grained context when extracting raw resources into triples~\cite{wang-han-2025-proprag}.
What remains missing is an evidence representation suited to predictive political reasoning, beyond treating external resources merely as knowledge-based evidence.

In this work, we identify two complementary signals encoded in political data. i) Actor stance signals, \textit{i.e.}, their subjective attitudes toward specific issues, which reflect their values, ideology, or partisan interests~\cite{Burnham_2025}. Although implicit in complex contexts, these signals possess greater reasoning potential compared to factual knowledge.
ii) High-order structure signals, \textit{i.e.}, the indirect dependencies within political actor interaction networks, which aim to characterize group influence.
Political actors are connected not only by direct ties but also through higher-order relational patterns. For example, an actor’s stance may influence others through intermediary organizations or decision-makers, forming an indirect network of policy influence. Such network-embedded dependencies are difficult to express in natural language, often leading to verbose descriptions and noise.
Semi-structured data formats provide a suitable basis for modeling these two signals, as they preserve contextual detail while retaining the interactions among political actors encoded in the structure.

In this work, we propose PSL, a dual-view framework that uses semi-structured political data to enhance LLMs for predictive political QA.
PSL extracts and integrates two complementary types of signals: i) actor stance signals from a semantic view and ii) high-order structure signals from a vector view. 
Specifically, PSL constructs an actor profile in JSON format for each political actor from abundant records, and links actors via shared records to build an interaction graph. 
In the semantic view, PSL retrieves question-relevant records from profiles and infers extensible stance signals to capture an actor’s preference on a specific issue.
In the vector view, PSL propagates information over the interaction graph to obtain actor representations that encode high-order neighborhood signals.
PSL then co-embeds the actor representation with the question-focus embedding, producing mutually refined vectors, and equips the LLM with structure-aware reasoning through lightweight fine-tuning.
Finally, PSL integrates stances from the semantic view with collaborative representations from the vector view and injects them into the LLM, providing stronger inferential evidence for predictive political QA.

This work makes three contributions.
First, we identify two complementary signals in semi-structured political records: actor stances and high-order structure signals. The former capture actors’ preferences on specific issues, whereas the latter characterize group influence revealed by shared political behavior.
Second, we propose PSL, a dual-view framework for these two signals: a semantic view extracts actor-stance signals, and a vector view models high-order structure signals, jointly enhancing LLMs’ predictive reasoning.
Third, we conduct comprehensive experiments on three real-world datasets across different LLMs. The results demonstrate that PSL significantly outperforms all the baseline methods.
Ablations verify the complementary benefits of semantic stance evidence and vectorized structure signals.

\section{The PSL Framework}

As illustrated in Figure~\ref{mainfig2}, PSL derives inference-friendly evidence from semi-structured data through a dual-view strategy. 
The semi-structured format provides foundational support for this design. Compared with KG triples, it preserves finer-grained context for stance extraction, while its structured fields support more precise retrieval and naturally enable the construction of an actor interaction graph.
On this basis, PSL models political evidence from two views.
Actor stances are expressed as natural-language judgments in the semantic view, whereas high-order structure signals are expressed as actor embeddings in the vector view.
To enable joint use of these complementary signals, PSL introduces two levels of synergy.
Within the vector view, inspired by collaborative filtering~\cite{schafer2007collaborative}, PSL co-embeds the actor and the question representations to produce mutually refined vectors, allowing the LLM to exploit structure signals in downstream reasoning.
Across the semantic and vector views, it injects stance and structure signals through a hybrid prompt.

\subsection{Actor Stance Acquisition}
\label{2.1}

\noindent\textbf{Political Behavior Profiles.}
We construct human-centered, domain-specific, and semi-structured political behavior profiles that encapsulate factual knowledge related to U.S. politics, aiming to facilitate political understanding without excessively compressing the underlying information.
We first collect politically relevant data sourced from legislative and diplomatic records. 
Specifically, we obtain legislative data via the LegiScan API, which contains legislators' voting records and bill descriptions. For diplomatic events, we extract interaction records involving political actors, particularly those pertinent to U.S. politics \cite{DVN/28075_2015}. 
We reorganize these raw data into a human-centered structure (\textit{i.e.}, $file(P)$). Based on these data, we extract a set of political actors, denoted by $P$. For each actor $p\in P$, we construct a corresponding political behavior profile, denoted by $file(p)$ in JSON format. 
The set of $file(p)$ is denoted as $file(P)$. 
Each profile $file(p)$ consists of a set of records, where each record $r_p \in file(p)$ is an object containing multiple attributes, and each attribute is expressed as $a_{r_p} \in r_p$, which is a key-value pair describing the actor’s political behavior. For legislative events, each record $r_p$ corresponds to a specific voting instance (\textit{i.e.}, voting record), with attributes such as the bill’s title, description, and the actor's vote.
For diplomatic events, $r_p$ denotes a concrete interaction event, with attributes detailing the action type and involved parties. 
We summarize bill descriptions with a summarization model~\cite{lewis2019bart} and add them as an additional attribute to reduce token consumption.

\noindent\textbf{Ladder Retrieval.}
\label{retrieval}
Given a question $q$, we aim to retrieve the most relevant records from $file(P)$. We first utilize an LLM to decompose $q$ into $p^*$ (the primary political actor) and $q^*$ 
(the question focus conveying the essential semantic content needed to infer an appropriate response~\cite{moldovan1999lasso}).
For instance, in ``What vote will the \{political actor\} cast on \{a description of a bill\}?”, $q^*$ corresponds to the bill description. This decomposition is formalized as $q \xrightarrow{\text{LLM}} (p^*,q^*).$
Once $p^*$ is obtained, we match it to its corresponding political actor $p_s$ and retrieve a set of highly relevant records from $file(p_s)$.
Since each $r_p$ contains multiple attributes, traditional embedding-based retrieval methods \cite{karpukhin2020dense,khattab2020colbert} often struggle to effectively exploit such structural attribute information and may introduce noise. 
Inspired by \cite{siddiquie2011image,macdonald2021approximate}, we propose ladder retrieval, an iterative retrieval strategy formalized as follows:
\begin{gather}
    R^{(i)} = LR(R^{(i-1)},q^*,k_{i}). 
    \label{eq:lr}
\end{gather}
This iterative process begins by initializing the record set $R^{(0)}$ as $file(p_s)$. During the $i$-th iteration, the retrieval function LR selects $k_i$ records from $R^{(i-1)}$ to generate $R^{(i)}$. 
The selection is based on the semantic similarity between $q^*$ and the $i$-th attribute of each record in $R^{(i-1)}$. 
After $n$ iterations, this process yields $R^{(n)}$, the top $k_n$ records most relevant to $q$.

\begin{figure}[t]
    \centering
    \includegraphics[width=0.88\linewidth]{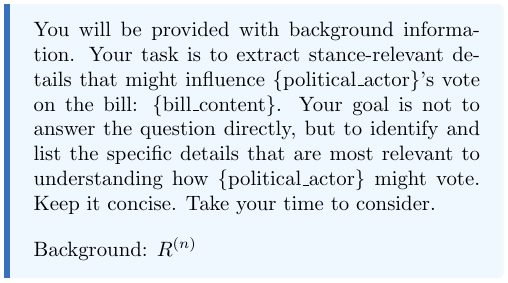}
    \caption{The prompt template of stance inference for distillation.}
    \label{fig:takeaway-example}
\end{figure}

\noindent\textbf{Stance Inference.} 
Factual political records provide indirect clues rather than explicit answers. We therefore extract their implicit preferences as concise, readable stance evidence.
Compared with raw facts, such evidence is more issue-aligned and exposes cues such as behavioral consistency and ideological orientation, providing a more targeted basis for LLM reasoning.
To reduce overhead, we distill a large teacher model into a SLM via task-specific knowledge distillation~\cite{gou2021knowledge}.
We construct political questions from behavior profiles by casting each record as a question, then retrieve the associated record set $R^{(n)}$ using ladder retrieval (Formula~\eqref{eq:lr}). We prompt the teacher to extract stance-relevant information from $R^{(n)}$, yielding high-quality task data (template in Figure~\ref{fig:takeaway-example}). 
During training, the SLM takes the same inputs as the teacher and is optimized with a cross-entropy loss between its logits and the teacher’s tokenized outputs, aligning its stance extraction behavior.
Thus, $q$ and $R^{(n)}$ can be input into the trained SLM via a predefined template as follows:
\begin{gather}
    R^{(n)}, q \xrightarrow{\text{SLM}} \mathcal{S},
\end{gather}
where $\mathcal{S}$ is the question-related stance (detailed in Appendix).

\subsection{High-Order Structure Signals Acquisition}
\label{2.2}

\noindent\textbf{Actor Interaction Graph.}
Although political behavior profiles are independent, their actor-centered semi-structured format naturally supports an interaction graph, where political actors are linked by shared records. This allows PSL to mine structural influence implicit in the data.
Specifically, we construct a node for each actor and record. For $r_p \in file(p)$, we construct an edge between the actor $p$ and the record $r_p$. 
We represent the resulting interaction graph as $\mathcal{G_I} = \{(p, e_{pr}, r)\} \subseteq P \times E \times R$, where $P$ is the actor set, $R$ is the record set, and $E$ is the set of edges representing interactions between actors and records. 
For each edge $e_{pr} \in E$ connecting $p$ and $r$, its weight is defined as follows: for a voting record, the weight is set to $1$ if $p$ voted ``\textit{Yea}'' on $r$, and $-1$ otherwise.
For a diplomatic record, the weight is set to $1$, indicating the existence of an interaction between $p$ and $r$. In this case, the absence of an interaction is denoted by the non-existence of $e_{pr}$ rather than being assigned a weight of $-1$.
For example, $p_1 \xrightarrow{e_{p_1 r_5}} r_5 \xleftarrow{e_{p_3 r_5}} p_3$ shows how a shared record connects two actors.

\noindent\textbf{Neighbor-Aware Actor Representation Construction.}
\label{embedding}
To obtain actor representations that capture structural influence, we propagate information over $\mathcal{G}_I$. Similar to prior work~\cite{he2020lightgcn}, our main goal is to extract structure signals from $\mathcal{G}_I$. We therefore adopt a simple weighted-sum aggregator that iteratively smooths node embeddings over the graph:
\begin{equation}
    \begin{gathered}
    \mathbf{e}_p^{(i+1)} = \sum _{r \in \mathcal{N}_p} e_{pr} \frac{1}{\sqrt{|\mathcal{N}_p||\mathcal{N}_r|}} \mathbf{e}_r^{(i)}, \\
    \mathbf{e}_r^{(i+1)} = \sum _{p \in \mathcal{N}_r} e_{pr} \frac{1}{\sqrt{|\mathcal{N}_p||\mathcal{N}_r|}} \mathbf{e}_p^{(i)}.
    \end{gathered}
    \label{eq:1}
\end{equation}
Here, the symmetric normalization term $1/{\sqrt{|\mathcal{N}_p||\mathcal{N}_r|}}$ follows the standard GCN design \cite{kipf2016semi}, where $\mathcal{N}_p$ (resp. $\mathcal{N}_r$) denotes the first-hop neighbors of $p$ (resp. $r$) and $\left| \mathcal{N}_p \right|$ (resp. $\left| \mathcal{N}_r \right|$) denotes the size of $\mathcal{N}_p$ (resp. $\mathcal{N}_r$). For each record $r$, we initialize its embedding as $\mathbf{e}_r$ via a frozen pre-trained embedding model \cite{reimers2019sentence}. 
The initial embeddings of political actors can be calculated by aggregating their neighbors as follows:
\begin{equation}
    \begin{gathered}
        \mathbf{e}_r^{(0)} = \mathbf{e}_r , \\
        \mathbf{e}_p^{(0)} = \sum _{r \in \mathcal{N}_p} e_{pr} \frac{1}{\sqrt{|\mathcal{N}_p|}\sqrt{|\mathcal{N}_r|}} \mathbf{e}_r.
    \end{gathered}
    \label{eq:2}
\end{equation}
After $l$ propagation layers, we obtain multi-layer representations of $p$, denoted as $(\mathbf{e}_p^{(0)},\mathbf{e}_p^{(1)},\dots,\mathbf{e}_p^{(l)})$. 
Inspired by \cite{xu2018representation}, we adopt a layer-aggregation mechanism to generate the political actor's final representation by concatenating the first and last embeddings as $\mathbf{\hat{e}}_p = \mathbf{e}_p^{(0)} || \mathbf{e}_p^{(l)}$, where $||$ denotes the concatenation operation.

\subsection{Synergistic Enhancement of LLM Reasoning}

\label{2.3}

To enable joint use of actor-stance and high-order structure signals, PSL introduces two levels of synergy. 
Within the vector view, actor representations reside outside the LLM’s language space and are difficult to use directly. 
Inspired by prior work showing that LLMs can exploit collaborative vectors~\cite{schafer2007collaborative,zhang2024text},
we innovatively treat the actor as a user-like entity and the question focus as an item-like entity, and model their interaction through the actor and question-focus embeddings.
The co-embedding module does more than reduce dimensionality into a shared latent space. By modeling actor–question interactions, it makes actor representations question-conditioned and acts as an implicit structural filter, highlighting relevant neighbors while suppressing structural noise.
Specifically, since the dimension of the corresponding actor embedding $\mathbf{\hat{e}}_{p_s}$ is twice that of the question focus embedding $\mathbf{e}_{q^*}$, we first align them by duplicating and concatenating $\mathbf{e}_{q^*}$ with itself, yielding $\mathbf{\hat{e}}_{q^*}  = \mathbf{e}_{q^*} || \mathbf{e}_{q^*}$ to ensure dimensional consistency.
Here, $q^*$ and $p_s$ are obtained from $\S$~\ref{2.1}, and $\mathbf{e}_{q^*}$ and $\mathbf{e}_{p_s}$ are their embeddings, respectively. The computation of $\mathbf{\hat{e}}_{p_s}$ is provided in $\S$~\ref{2.2}.   
Then, we feed $\mathbf{\hat{e}}_{p_s}$ and $\mathbf{\hat{e}}_{q^*}$ into the co-embedding module (\textit{i.e.}, a trained MLP) separately to generate their collaborative representation:
\begin{gather}
    \mathbf{\hat{e}}_{p_s}, \mathbf{\hat{e}}_{q^*} \xrightarrow{\text{MLP}}  
    \mathbf{\hat{e}}_{p_s}^*, \mathbf{\hat{e}}_{q^*}^* .
    \label{eq:mlp}
\end{gather}
This process enables the LLM to more efficiently leverage their collaborative relationship to capture and utilize the high-order structure signals encoded in the actor representation.

Next, to achieve synergy across the semantic and vector views, we integrate political actor stances with collaborative embeddings to support the LLM’s reasoning.
This is achieved by populating a predefined instruction template $T$ with both explicit and implicit signals, and then injecting the filled template into the LLM to generate final answers that are grounded in the associated external knowledge:
\begin{gather}
    T(q,\mathcal{S},\hat{\mathbf{e}}_{p_s}^*,\hat{\mathbf{e}}_{q^*}^*) \xrightarrow{\text{LLM}_{\Phi+\Phi'}} Answer ,
\end{gather}
where $\text{LLM}_{\Phi+\Phi'}$ denotes the LLM slightly fine-tuned with LoRA \cite{hu2022lora}. The tuning process will be detailed in the next section.

\begin{table*}[!t]
\centering
\small
\resizebox{\textwidth}{!}{%
\begin{tabular}{lrrrrrrrrrr}
\toprule
\multirow{2}{*}{\textit{\textbf{Method}}} & \multicolumn{3}{c}{\textit{\textbf{RCVP}}} & \multicolumn{3}{c}{\textit{\textbf{ICEWS}}} & \multicolumn{3}{c}{\textit{\textbf{StaId}}} \\
\cmidrule(lr){2-4} \cmidrule(lr){5-7} \cmidrule(lr){8-10}
                           & \textit{Llama-8B} & \textit{Mistral-7B} & \textit{Deepseek-7B} & \textit{Llama-8B} & \textit{Mistral-7B} & \textit{Deepseek-7B} & \textit{Llama-8B} & \textit{Mistral-7B} & \textit{Deepseek-7B} \\
\midrule
Vanilla                    & 27.36             & 27.33               & 28.03    & 18.42             & 19.18               & 19.67                & 35.00             & \underline{33.36}   & 29.59                \\
GKP (ACL'22)               & 23.07             & 26.29               & 24.28                & 7.09              & 14.37               & 9.65                 & \underline{38.26} & 33.04               & 34.76    \\
RECITE (ICLR'23)           & 25.08             & 23.38               & 14.90                & 14.89             & 16.95               & 11.81                & 31.88             & 29.47               & 27.49                \\
LangChain                  & 31.10             & 35.52   & 24.37                & 8.06              & 18.06               & 19.06                & 27.71             & 29.51               & 30.07                \\
InstructRAG (ICLR'25)      & 29.84             & 21.64               & 26.81                & 11.57             & 7.71                & 18.21                & 31.53             & 28.30               & 32.97                \\
KAPING (ACL'23)            & 23.15             & 23.96               & 23.59                & 20.52 & \underline{22.52}   & 15.36                & 36.20             & 30.28               & 34.22                \\
MindMap$_{route}$ (ACL'24) & 20.57             & 25.36               & 22.35                & 13.54             & 19.23               & 14.89                & 34.98             & 32.31               & 31.76                \\
MindMap$_{lang}$ (ACL'24)  & 21.37             & 24.19               & 21.91                & 12.69             & 19.04               & 11.61                & 35.69             & 31.51               & 32.51                \\
MindMap (ACL'24)           & 18.66             & 23.02               & 21.54                & 12.70             & 22.01               & 21.70    & 34.72             & 31.97               & 31.51                \\
PAA (AAAI'25)              & \underline{44.35} & \underline{41.78}                   & \underline{42.35}                    & \underline{31.35}                 & 18.99                   & \underline{24.53}                     & 36.34                 & 27.41                   & \underline{39.89}                     \\
PSL                        & \textbf{55.92}    & \textbf{50.30}      & \textbf{43.59}       & \textbf{57.81}    & \textbf{54.69}      & \textbf{53.42}       & \textbf{48.50}    & \textbf{48.83}      & \textbf{50.58}       \\
\bottomrule
\end{tabular}}
\caption{Experimental results of PSL with baselines. The best results are highlighted in bold, and the second best results are underlined.}
\label{effectiveness}
\end{table*}

\subsection{Learning}
\label{2.4}

\noindent\textbf{MLP Training.}
To co-embed neighbor-aware political actor representations with question representations, we leverage the records of political actors as $q^*$ to train the MLP defined by Formula \eqref{eq:mlp}.
We design an objective function based on the Bayesian personalized ranking loss, which encourages the model to assign higher prediction scores to observed (positive) records than to unobserved (negative) ones:
\begin{gather}
    \mathcal{L}\! =\! -\! \sum_{p \in P} \!\sum_{r^+ \in  \mathcal{N}_p^+}\! \sum_{r^- \in  \mathcal{N}_p^-} \! \ln \sigma( y_{_{pr^+}} - y_{_{pr^-}} )\! +\! \lambda \left\| \Theta \right\|^2 ,
\end{gather}
where $y_{_{pr}}= {(\hat{\mathbf{e}}^*_p)} ^{T}\hat{\mathbf{e}}_{r}^*$ denotes the inner product between ${\hat{\mathbf{e}}_p^*}$ and $\hat{\mathbf{e}}_{r}^*$,
$\lambda$ controls the $L_2$ regularization strength, 
$\Theta$ denotes the parameters of the MLP, 
$\mathcal{N}_p^+$ denotes the set of positive (observed) records, referring to those records for which the weight of $e_{pr}$ is 1,
$\mathcal{N}_p^-$ denotes the set of negative (unobserved) records, including explicit negative voting records (the weight of $e_{pr}$ is -1) and unobserved diplomatic records ($e_{pr}$ does not exist). More details are provided in Appendix. 

\noindent\textbf{LoRA Tuning.} 
We randomly select $800$ pairs of $\hat{\mathbf{e}}_p$ and $\hat{\mathbf{e}}_r$, where $\hat{\mathbf{e}}_r$ is uniformly sampled from $\mathcal{N}_p^+$ and $\mathcal{N}_p^-$. 
These pairs are processed by the trained MLP to produce collaborative embeddings $\hat{\mathbf{e}}_p^*$ and $\hat{\mathbf{e}}_r^*$. 
Based on collaborative embeddings, we construct instructional samples via the natural language concatenation of $\hat{\mathbf{e}}_p^*$ and $\hat{\mathbf{e}}_r^*$ as input, and the relation between $p$ and $r$, that is, the corresponding relation label $e_{pr}$ as output.
These instructional samples are then used to fine-tune the LLM using LoRA (detailed in Appendix).

\section{Experiments}
\subsection{Experimental Setting}

\noindent\textbf{Datasets.}
We conduct experiments on three available datasets \cite{mou2024unifying} covering different political scenarios: (1) RCVP; (2) ICEWS; (3) StaId.
The source code and datasets used in our paper are publicly available\footnote{https://github.com/zhouzihan-liu/PSL}.

\noindent\textbf{Evaluation Metrics.}
Following \cite{mou2024unifying}, we adopt the macro F1 score as the evaluation metric for the binary classification tasks (\textit{i.e.}, RCVP and StaId). For the multiple-choice task in ICEWS, we report accuracy.

\noindent\textbf{Setting Details.}
\label{setting}
To assess the effectiveness of PSL, we employ 
Llama-3.1-8B-Instruct~\cite{grattafiori2024llama3herdmodels}, 
Mistral-7B-Instruct~\cite{jiang2023mistral},
Deepseek-7B-Chat~\cite{bi2024deepseek},
and GPT-3.5-Turbo\footnote{https://platform.openai.com/docs/models/gpt-3.5-turbo} as backbone models, with all experiments conducted under this default configuration unless otherwise noted.
In the distillation stage, Flan-T5-Small~\cite{chung2024scaling} is used as the student SLM, while GPT-4o-mini\footnote{https://platform.openai.com/docs/models/gpt-4o-mini} serves as the teacher model.
The number of iterations in the ladder retrieval process, retrieved candidates $k_n$, and propagation layers $l$ are set to $3$, $5$, and $2$, respectively.
The data used to train the MLP is entirely from the profile set. The network architecture of the MLP is configured as $1536 \xrightarrow{} 1024\xrightarrow{} 512\xrightarrow{} 256\xrightarrow{} 64\xrightarrow{} 32$. 
The regularization coefficient $\lambda$ is set to $1 \times 10^{-4}$, and dropout is applied during training to improve generalization.
Following \cite{zheng2024llamafactory}, we fine-tune Llama using $800$ samples for $3$ epochs, with all other hyperparameters kept at their default settings.

\subsection{Effectiveness Study} 
We compare PSL with the following $12$ methods. 
Vanilla, \textit{i.e.}, directly inputs questions into the LLM without any additional knowledge. GKP \cite{liu2022generated} prompts the LLM to generate knowledge relevant to the question and then appends these generated statements into the prompt to improve reasoning. RECITE \cite{sun2023recitationaugmented} prompts the LLM to ``recite” self‐sampled memory‐like passages before answering, leveraging internalized knowledge for better QA performance. LangChain \cite{Chase_LangChain_2022} retrieves political documents based on semantic similarity and uses them as prompts to improve answer generation. InstructRAG \cite{wei2025instructrag} prompts the LLM to generate reasoning steps before answering, helping it better use the retrieved political documents. KAPING \cite{baek2023knowledge} retrieves political knowledge from KG based on semantic similarity and formats them as triples in prompts. MindMap \cite{wen2024mindmap} generates both the answer and a reasoning path in a tree-like format for interpretability. MindMap$_{route}$ structures retrieved triples into paths, such as ``climate policy debate $\xrightarrow{}$ related politicians $\xrightarrow{}$ John Smith, Emily Carter", and uses them to guide reasoning. MindMap$_{lang}$ converts the structured knowledge paths into natural language narratives, which are then used to prompt the LLM. PEG$_{exp\_sum}$ \cite{mou2024unifying} retrieves relevant political triples and summarizes them using the LLM to facilitate downstream reasoning. PEG$_{exp\_GTR}$ \cite{mou2024unifying} clusters and summarizes selected triples via LLMs for final answer generation. 
PAA \cite{li2025political} utilizes political actor profiles to guide LLMs in simulating legislative behavior for roll-call vote prediction. 
For document-enhanced methods (\textit{i.e.}, LangChain and InstructRAG), we treat each record in profiles as an independent document. For KG-enhanced methods (\textit{i.e.}, KAPING, MindMap$_{route}$, MindMap$_{lang}$, MindMap, PEG$_{exp\_sum}$, and PEG$_{exp\_GTR}$), the political KG MVPKG \cite{mou2024unifying} is uniformly adopted as the external knowledge source. For the agent-based method PAA, we follow the original approach and extract $20$ records per political actor from our constructed profiles. 
Note that the raw data used to construct MVPKG fully encompasses the profiles we construct.
To report the comparative results with PEG, since its core implementation has not been publicly released, we present the performance metrics as reported in the original paper. The results are summarized in Table~\ref{effectiveness--}.

From the results shown in Table \ref{effectiveness}, we can see that PSL outperforms all baselines on three datasets with different LLMs. Models without external knowledge (\textit{i.e.}, GKP and RECITE) usually fail to surpass Vanilla, as LLMs struggle to generate accurate knowledge about political facts or future events. 
Document-enhanced methods perform well only over RCVP, with limited effectiveness elsewhere, due to challenges in accurately retrieving relevant information and the interference caused by direct text injection in complex reasoning over ICEWS and StaId. In contrast, KG-enhanced methods show stronger and more consistent performance. KAPING remains stable across three datasets, while MindMap and its variants, relying on entity-path retrieval, are less effective in future-oriented scenarios. 
PAA achieves notable performance over RCVP. 
Although both PEG and PAA are designed to enhance LLMs in political tasks using external knowledge similar in content to that employed by PSL, PSL consistently outperforms them in all experimental setups, which may be attributed to the fact that PSL can mine political knowledge from semi-structured data to enhance the LLM. Furthermore, PSL also demonstrates competitive time efficiency, as detailed in Appendix.

\begin{table}[!t]
\centering
\resizebox{0.78\linewidth}{!}{%
\begin{tabular}{lrrr}
\toprule
\multirow{1}{*}{\textit{\textbf{Method}}} & \multicolumn{1}{c}{\textit{\textbf{RCVP}}}      
& \multicolumn{1}{c}{\textit{\textbf{ICEWS}}}   
& \multicolumn{1}{c}{\textit{\textbf{StaId}}}   \\ 
                            \midrule   
Vanilla                     & 34.11               & 19.40              & 33.24             \\
GKP (ACL'22)                & 20.43               & 15.40              & 45.32             \\
RECITE (ICLR'23)            & 15.79               & 21.40              & 35.57             \\
KAPING (ACL'23)             & 37.83               & 20.80              & 36.99             \\
MindMap$_{route}$ (ACL'24)  & 32.60               & 28.00              & 35.03             \\
MindMap$_{lang}$ (ACL'24)   & 38.57               & 28.40              & 42.19             \\
MindMap (ACL'24)            & 38.72               & 25.60              & 23.38             \\
PEG$_{exp\_sum}$ (WWW'24)   & 40.62               & 26.40              & 42.12             \\
PEG$_{exp\_GTR}$ (WWW'24)   & \underline{41.21}               & \underline{28.60}  & \underline{46.21}             \\
PSL                         & \textbf{50.23}      & \textbf{45.89}     & \textbf{49.90}   \\
                            \bottomrule
\end{tabular} 
}

\caption{Experimental results of PSL with all baselines under GPT-3.5. 
The best results are highlighted in bold, and the second best results are underlined. 
Results are taken from \cite{mou2024unifying}, except for PSL.}   

\label{effectiveness--}

\end{table}

\begin{table}[!b]
\centering
\small
\resizebox{0.88\linewidth}{!}{%
\begin{tabular}{lrrr}
\toprule
\textit{\textbf{Ablations}}         & \textit{\textbf{RCVP}}  & \textit{\textbf{ICEWS}} & \textit{\textbf{StaId}} \\
\midrule
PSL                      & \textbf{55.92} & \textbf{57.81} & \textbf{48.50} \\ 
w/o Actor Stance Acquisition              & 46.15          & 39.07          & 46.17          \\
w/o High-Order Structure  & 39.35          & 49.02          & 46.35          \\
\hspace{1.56em} Signals Acquisition  & & & \\
\bottomrule
\end{tabular}
}
\caption{The results of ablation study for PSL.}
\label{ablation}
\end{table}

\subsection{Ablation Study}

We validate PSL's components by removing actor stance (\textit{w/o Stance}) or high-order structure (\textit{w/o Structure}) on Llama-8B. Note that \textit{w/o Structure} eliminates collaborative embeddings, restricting inputs to stance and question text. Results in Table \ref{ablation} show: (1) \textit{w/o Stance} causes significant drops across all datasets, validating that retrieved records indeed capture critical issue-specific positions; (2) \textit{w/o Structure} degrades performance, notably on RCVP and ICEWS , confirming that high-order signals capture group-level behavioral dependencies essential for voting and diplomatic prediction.

\subsection{Parameter Study}

\begin{figure}[b] 
   \centering
    \begin{subfigure}{0.49\linewidth}
        \centering
        \includegraphics[width=\linewidth]{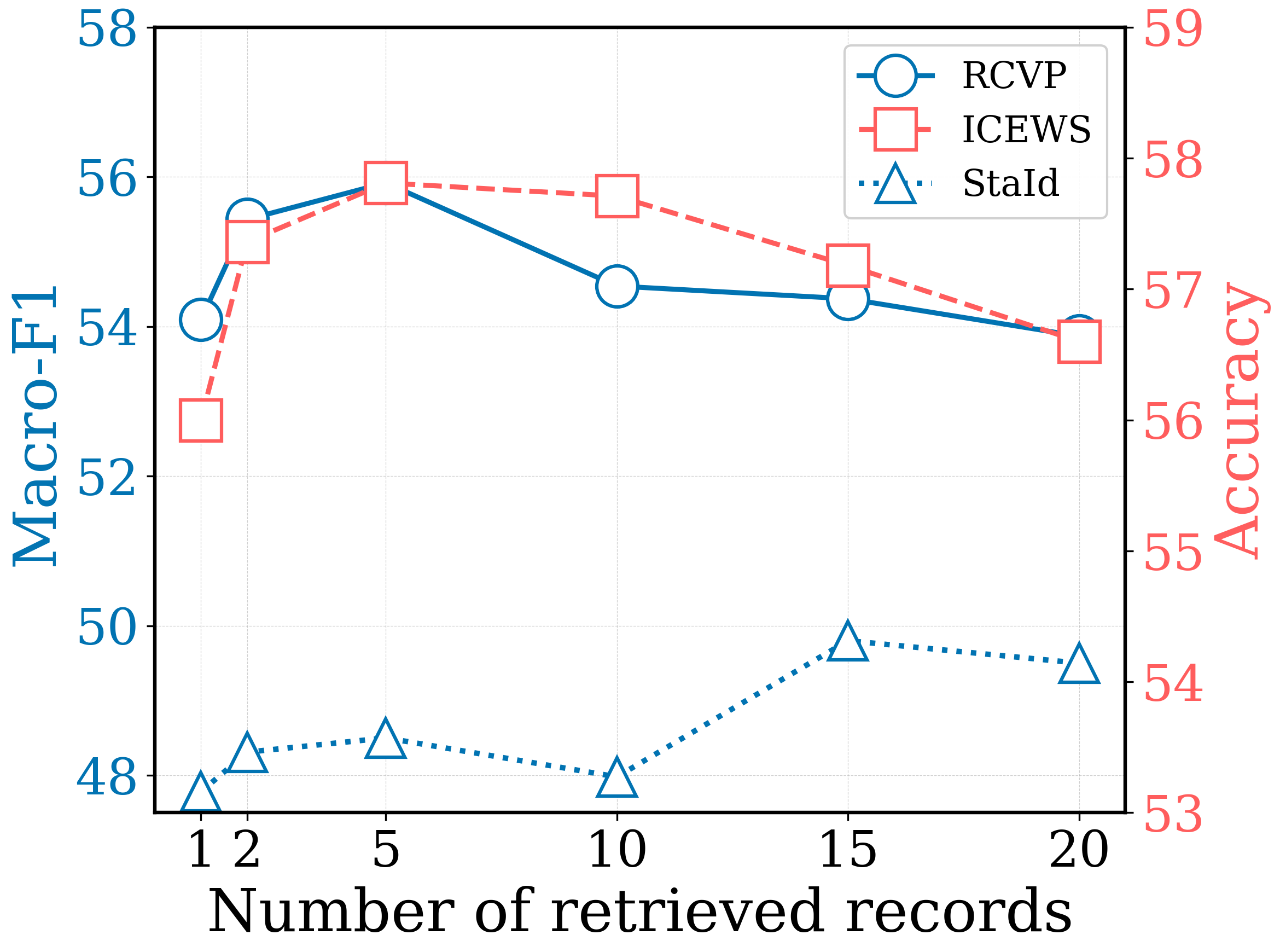}
        \caption{With varied parameter of the retrieved record number.}
    \end{subfigure}
    \hfill
    \begin{subfigure}{0.49\linewidth}
        \centering
        \includegraphics[width=\linewidth]{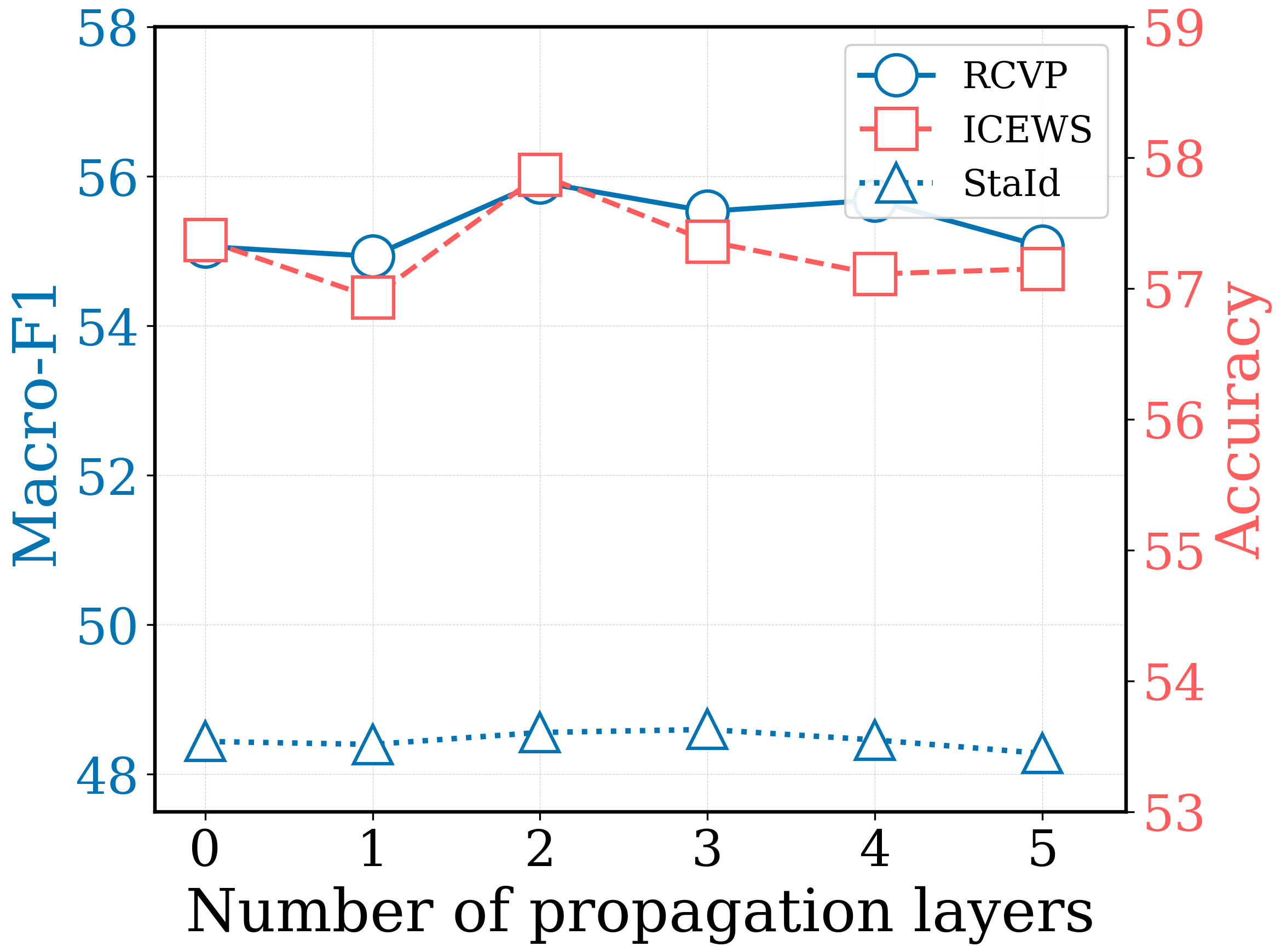}
        \caption{With varied parameter of the propagation layer number.}
    \end{subfigure}

    \caption{Parameter study on the RCVP, ICEWS, and StaId datasets.}
    \label{parameter}
\end{figure}

We examine the impact of retrieved records and propagation layers using Llama-8B (Figure \ref{parameter}). (1) Retrieved Records: Performance improves initially, peaking at 5 records for RCVP/ICEWS and 15 for StaId, before declining due to noise from irrelevant content. StaId shows a wider growth range, suggesting a need for richer context. Notably, PSL surpasses baselines across all settings. (2) Propagation Layers: Performance peaks at layer 2 (RCVP/ICEWS) and layer 3 (StaId). This confirms that limited propagation captures essential structure signals, whereas deeper layers lead to over-smoothing and reduced discriminative power.

\section{Further Analysis}

\noindent\textbf{Effect Analysis of Different Components of Actor Stance Acquisition.}
We investigate the impact of retrieval and stance inference by evaluating PSL variants (without stance inference) using TF-IDF~\cite{aizawa2003information}, BM25~\cite{robertson2009probabilistic}, Embedding~\cite{reimers2019sentence}, and our Ladder Retrieval (PSL$_{LR}$). 
As shown in Table~\ref{p1}, all variants surpass the baselines and the ``w/o Stance'' ablation, demonstrating that profile records provide critical context for prediction. 
Specifically, lexical methods (BM25, TF-IDF) yield lower scores due to their reliance on lexical overlap, failing to capture semantic or structural dependencies. 
Conversely, PSL$_{LR}$ achieves the best performance among retrieval variants. This validates LR's effectiveness in leveraging hierarchical structural information and handling missing data through score inheritance. 
Ultimately, the full PSL framework outperforms all variants, confirming that transforming isolated records into targeted stance information reduces noise and boosts overall performance.

\begin{figure}[!t]
  \centering
  \begin{minipage}{\linewidth} 
    \centering

  \begin{subfigure}[b]{\linewidth}
    \centering
    \includegraphics[width=0.9\linewidth]{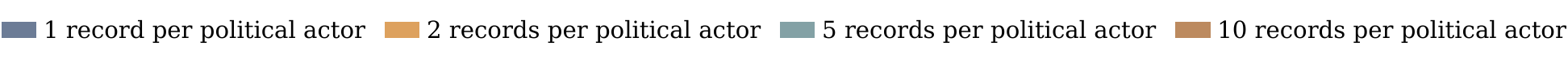}
  \end{subfigure}

  \begin{subfigure}[b]{0.332\linewidth}
    \centering
    \includegraphics[width=\linewidth]{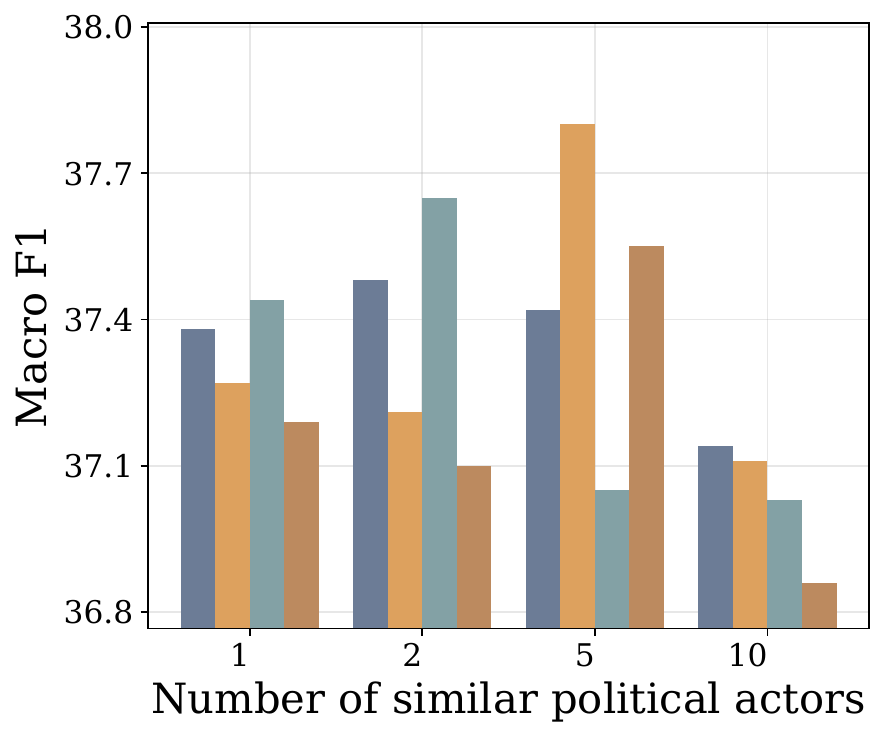}
    \caption{RCVP}
  \end{subfigure}\hfill
  \begin{subfigure}[b]{0.332\linewidth}
    \centering
    \includegraphics[width=\linewidth]{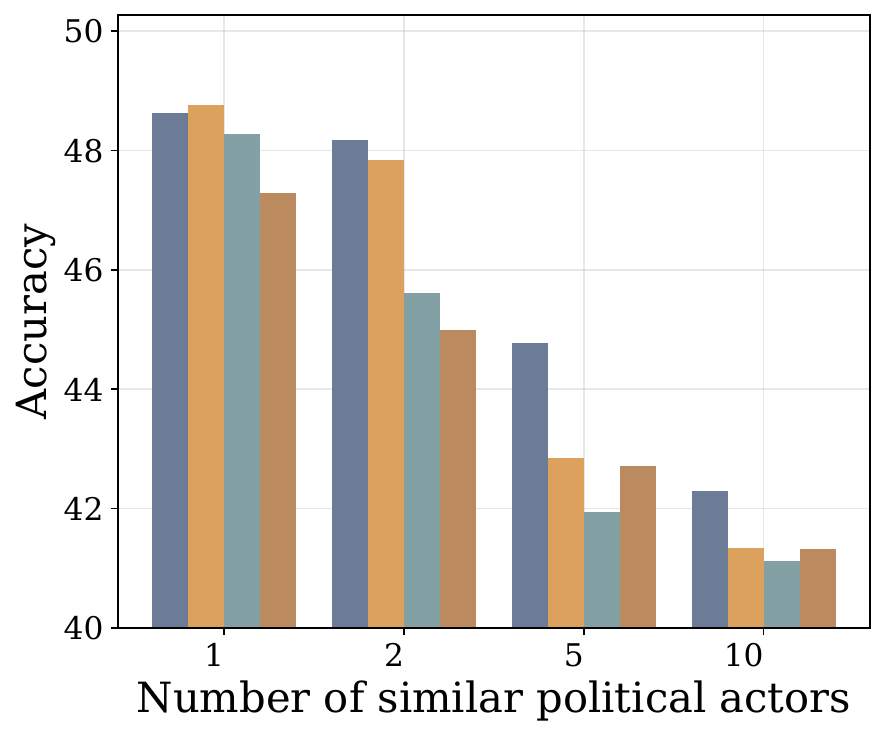}
    \caption{ICEWS}
  \end{subfigure}\hfill
  \begin{subfigure}[b]{0.332\linewidth}
    \centering
    \includegraphics[width=\linewidth]{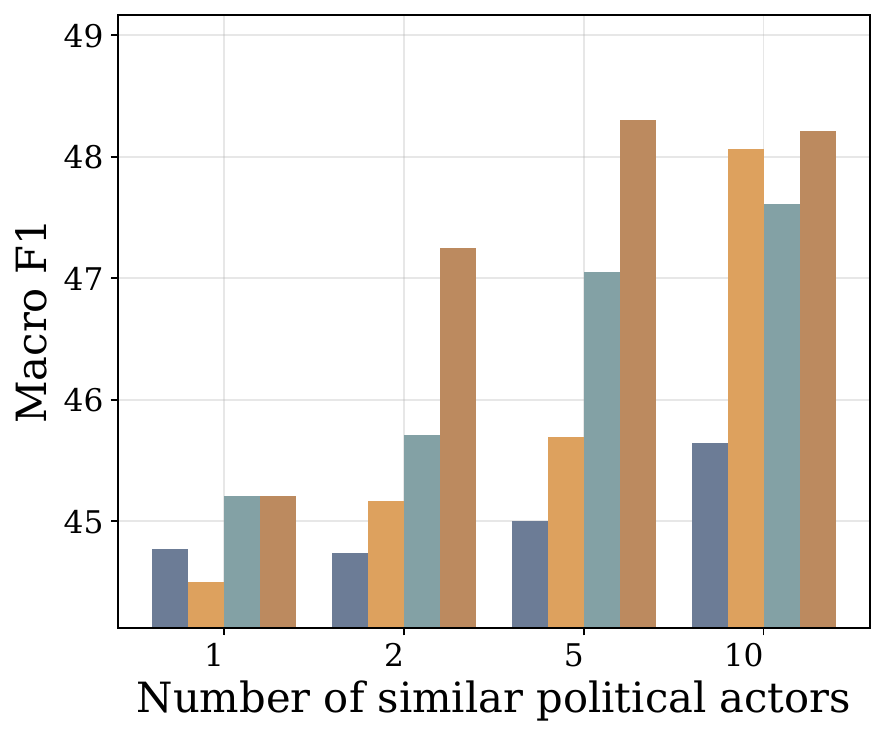}
    \caption{StaId}
  \end{subfigure}
  
  \end{minipage}

  \caption{Performance of PSL using different numbers of political actors’ neighbors and their records in the form of text.}
  \label{p2}
\end{figure}

\noindent\textbf{Effect Analysis of The Presentation Form for High-Order Structure Signals.}
We investigate representing high-order structure as text by retrieving records from similar neighbors. For each actor, we rank its directly connected neighbors by this similarity and select the top ones. Similarity between two actors is computed from their shared record links: for each record connected to both, we add 1 if their edge weights match (both +1 or both -1) and subtract 1 if they differ.
Figure~\ref{p2} shows the impacts.
RCVP benefits from distributing retrieval across more actors, as top-ranked neighbors provide the most valuable signals.
Conversely, ICEWS performance declines with increased context due to noise from indirect records, whereas StaId improves with retrieval volume to compensate for missing statement profiles.
Crucially, all text-based settings consistently underperform PSL.
This corroborates our assertion that articulating complex relations in natural language leads to information overload, validating the necessity of PSL's collaborative vector modeling.

\begin{table}[t]
\centering
\small
\resizebox{0.78\linewidth}{!}{%
\begin{tabular}{lrrr}
\toprule
\textit{\textbf{Variant}}         & \textit{\textbf{RCVP}}  & \textit{\textbf{ICEWS}} & \textit{\textbf{StaId}} \\
\midrule
PSL                      & \textbf{55.92} & \textbf{57.81} & \textbf{48.50} \\ 
PSL$_{TF-IDF}$  w/o SI                  & 52.07          & 56.70          & 46.45          \\
PSL$_{BM25}$   w/o SI                 & 51.87          & 57.22          & 45.94          \\
PSL$_{Embedding}$  w/o SI             & 52.96          & 57.16          & 46.69          \\
PSL$_{LR}$   w/o SI                    & 53.81          & 57.34          & 47.10          \\
\bottomrule
\end{tabular}
}
\caption{Performance of different variants of PSL. SI denotes the procedure of stance inference.}
\label{p1}
\end{table}

\section{Related Work}

\noindent\textbf{LLM Augmentation via External Knowledge.}
External knowledge augmentation grounds LLMs in resources beyond their parameters. Document-based methods, such as REALM \cite{guu2020retrieval}, RAG \cite{lewis2020retrieval}, and RePlug \cite{shi2024replug}, retrieve relevant passages, whereas KG-based methods inject compact factual triples. KAPING \cite{baek2023knowledge} and DiagLink \cite{zhou2026diaglink} uses KG triples for question answering; GPT4Graph \cite{guo2023gpt4graph} and NLGraph \cite{wang2023can} study graph reasoning with LLMs; and ToG \cite{sun2024thinkongraph} and MindMap \cite{wen2024mindmap} construct reasoning paths over graph structure. These methods are mainly designed for settings where external knowledge can provide answers or explicit reasoning paths. Predictive political QA is different: external resources record past behavior rather than the target behavior itself.
Recent work has adapted external augmentation to political reasoning. PEG \cite{mou2024unifying} injects retrieved triples from a political KG, whereas PAA \cite{li2025political} simulates political actors from textual profiles. Both improve political reasoning, but their evidence forms are limited for prediction: KG triples lose fine-grained context, and textual profiles weakly capture implicit group influence. We instead derive stance evidence and structure signals from semi-structured political records.

\noindent\textbf{Political Actor Modeling.}
Political actor modeling is central to computational political science. Classical approaches estimate actors’ ideological positions from roll-call votes, most notably through ideal point models \cite{clinton2004statistical}, later extended with legislative text \cite{gerrish2011predicting,kraft2016embedding}. Graph-based methods further model political relations through co-sponsorship networks \cite{yang2021joint}, contribution networks \cite{davoodi2020understanding}, and donation networks \cite{davoodi2022modeling}. Recent work enriches actor representations with external knowledge, including media coverage \cite{feng2021kgap}, social media statements \cite{mou2021align}, Wikipedia, and policy-research materials \cite{feng2022political}.
Despite their effectiveness, the heterogeneity of these data sources results in high collection costs.
In this paper, we construct political behavior profiles for political actors using open-access and continuously updated sources and inject actors' stances and their complex political relationships into LLMs.

\section{Conclusion}

We propose PSL, a framework that transforms semi-structured political records into inference-oriented evidence for predictive political QA. PSL enables textual semantics and graph-structural features to jointly strengthen LLM reasoning.
This dual-view design, grounded in semi-structured data, provides a more prediction-oriented evidence representation. Moreover, compared with KG triples, PSL better preserves contextual detail; compared with text-profile simulation, it more fully models structural relations among actors.
Experiments on multiple real-world datasets and LLMs show that PSL consistently outperforms existing augmentation methods. More broadly, our results provide initial evidence that inference-oriented evidence beyond knowledge-based evidence is effective for predictive political QA. We will release the code and model checkpoints to facilitate further research.

\section*{Limitations}
\textbf{Data scope.} This paper validates PSL in U.S. political settings, using public records such as legislative votes, bill information, and diplomatic events to construct semi-structured profiles and an interaction graph. This setting provides a clear experimental basis for testing PSL’s core design: extracting actor stance signals and high-order structure signals from commonly available political records to enhance predictive political QA. 
However, there remains room to broaden the data scope. 
On the one hand, future work could apply PSL to more countries, political systems, and language contexts to assess its generalizability across political environments. 
On the other hand, existing records could also be extended to richer sources, such as co-sponsorship, committee activities, public statements, campaign donations, and policy texts. These extensions do not require changes to PSL’s basic design, but could provide more comprehensive behavioral evidence for stance extraction and structure modeling.

\noindent\textbf{Signal fusion.} This paper designs two complementary signals to enhance predictive political QA. Experiments and ablation studies show that these signals provide effective reasoning evidence for LLMs from different perspectives. However, different questions may rely on the two signals to different degrees. Some questions may depend more on an actor’s historical stance on related issues, whereas others may rely more on group-level behavioral patterns among political actors. Future work could explore adaptive fusion mechanisms that dynamically adjust the weights of the two signals according to the question, actor, or political setting, enabling finer-grained modeling of evidence needs across predictive tasks.

\section*{Acknowledgments}
The work was partially supported by the National Key Research and Development Program of China (No. 2024YFF0617702); the National Natural Science Foundation of China (Nos. U22A2025, 62402097, 62232007, and U23A20309); the Joint Funds of the Natural Science Foundation of Liaoning Province (No. 2023-BSBA-132); the 111 Project (No. B16009); the Ant Group Research Program (No. 2025021900003); and the Fundamental Research Funds for the Central Universities (No. N2417007)

\bibliography{custom}

@misc{DVN/28075_2015,
author = {Boschee, Elizabeth and Lautenschlager, Jennifer and O'Brien, Sean and Shellman, Steve and Starz, James and Ward, Michael},
publisher = {Harvard Dataverse},
title = {{ICEWS Coded Event Data}},
UNF = {UNF:6:NOSHB7wyt0SQ8sMg7+w38w==},
year = {2015},
version = {V37},
doi = {10.7910/DVN/28075},
url = {https://doi.org/10.7910/DVN/28075}
}

@misc{Chase_LangChain_2022,
author = {Chase, Harrison},
month = oct,
title = {{LangChain}},
url = {https://github.com/langchain-ai/langchain},
year = {2022}
}

@inproceedings{zheng2024llamafactory,
  author = {Yaowei Zheng and Richong Zhang and Junhao Zhang and Yanhan Ye and Zheyan Luo and Zhangchi Feng and Yongqiang Ma},
  booktitle = {ACL},
  pages = {400--410},
  title = {LlamaFactory: Unified Efficient Fine-Tuning of 100+ Language Models},
  year = {2024}
}

@inproceedings{kipf2016semi,
  author       = {Thomas N. Kipf and
                  Max Welling},
  title        = {Semi-Supervised Classification with Graph Convolutional Networks},
  booktitle    = {ICLR},
  year         = {2017},
  pages        = {2713--2726}
}

@inproceedings{rag1,
  author = {Florin Cuconasu and
    Giovanni Trappolini and
    Federico Siciliano and
    Simone Filice and
    Cesare Campagnano and
    Yoelle Maarek and
    Nicola Tonellotto and
    Fabrizio Silvestri},
  booktitle = {SIGIR},
  pages = {719--729},
  title = {The Power of Noise: Redefining Retrieval for {RAG} Systems},
  year = {2024}
}

@inproceedings{yang2021joint,
  author = {Yuqiao Yang and
    Xiaoqiang Lin and
    Geng Lin and
    Zengfeng Huang and
    Changjian Jiang and
    Zhongyu Wei},
  booktitle = {IJCAI},
  pages = {1424--1430},
  title = {Joint Representation Learning of Legislator and Legislation for Roll
Call Prediction},
  year = {2020}
}

@inproceedings{galla2018predicting,
  author = {Divyanshi Galla and
    James Burke},
  booktitle = {MLDM},
  pages = {103--116},
  title = {Predicting Social Unrest Using {GDELT}},
  year = {2018}
}

@inproceedings{wen2024mindmap,
  author = {Yilin Wen and
    Zifeng Wang and
    Jimeng Sun},
  booktitle = {ACL},
  pages = {10370--10388},
  title = {MindMap: Knowledge Graph Prompting Sparks Graph of Thoughts in Large
Language Models},
  year = {2024}
}

@inproceedings{mou2024unifying,
  author = {Xinyi Mou and
    Zejun Li and
    Hanjia Lyu and
    Jiebo Luo and
    Zhongyu Wei},
  booktitle = {WWW},
  pages = {2603--2614},
  title = {Unifying Local and Global Knowledge: Empowering Large Language Models
as Political Experts with Knowledge Graphs},
  year = {2024}
}

@inproceedings{li2025political,
  author = {Hao Li and
    Ruoyuan Gong and
    Hao Jiang},
  booktitle = {AAAI},
  pages = {388--396},
  title = {Political Actor Agent: Simulating Legislative System for Roll Call
Votes Prediction with Large Language Models},
  year = {2025}
}

@inproceedings{moldovan1999lasso,
  author = {Dan I. Moldovan and
    Sanda M. Harabagiu and
    Marius Pasca and
    Rada Mihalcea and
    Richard Goodrum and
    Roxana Girju and
    Vasile Rus},
  booktitle = {TREC},
  pages={65--73},
  title = {{LASSO:} {A} Tool for Surfing the Answer Net},
  year = {1999}
}

@inproceedings{karpukhin2020dense,
  author={Karpukhin, Vladimir and Oguz, Barlas and Min, Sewon and Lewis, Patrick SH and Wu, Ledell and Edunov, Sergey and Chen, Danqi and Yih, Wen-tau},
  booktitle = {EMNLP},
  pages = {6769--6781},
  title = {Dense Passage Retrieval for Open-Domain Question Answering},
  year = {2020}
}

@inproceedings{khattab2020colbert,
  author = {Omar Khattab and
    Matei Zaharia},
  booktitle = {SIGIR},
  pages = {39--48},
  title = {ColBERT: Efficient and Effective Passage Search via Contextualized
Late Interaction over {BERT}},
  year = {2020}
}

@inproceedings{xu2018representation,
  author = {Keyulu Xu and
    Chengtao Li and
    Yonglong Tian and
    Tomohiro Sonobe and
    Ken{-}ichi Kawarabayashi and
    Stefanie Jegelka},
  booktitle = {ICML},
  pages = {5449--5458},
  title = {Representation Learning on Graphs with Jumping Knowledge Networks},
  year = {2018}
}

@inproceedings{guu2020retrieval,
  author = {Kelvin Guu and
    Kenton Lee and
    Zora Tung and
    Panupong Pasupat and
    Ming{-}Wei Chang},
  booktitle = {ICML},
  pages = {3929--3938},
  title = {Retrieval Augmented Language Model Pre-Training},
  year = {2020}
}

@inproceedings{shi2024replug,
  author = {Weijia Shi and
    Sewon Min and
    Michihiro Yasunaga and
    Minjoon Seo and
    Richard James and
    Mike Lewis and
    Luke Zettlemoyer and
    Wen{-}tau Yih},
  booktitle = {NAACL},
  pages = {8371--8384},
  title = {{REPLUG:} Retrieval-Augmented Black-Box Language Models},
  year = {2024}
}

@inproceedings{sun2024thinkongraph,
  author = {Jiashuo Sun and
    Chengjin Xu and
    Lumingyuan Tang and
    Saizhuo Wang and
    Chen Lin and
    Yeyun Gong and
    Lionel M. Ni and
    Heung{-}Yeung Shum and
    Jian Guo},
  booktitle = {ICLR},
  pages = {14199--14229},
  title = {Think-on-Graph: Deep and Responsible Reasoning of Large Language Model
on Knowledge Graph},
  year = {2024}
}

@inproceedings{siddiquie2011image,
  author = {Behjat Siddiquie and
    Rog{\'{e}}rio Schmidt Feris and
    Larry S. Davis},
  booktitle = {CVPR},
  pages = {801--808},
  title = {Image ranking and retrieval based on multi-attribute queries},
  year = {2011}
}

@inproceedings{macdonald2021approximate,
  author = {Craig Macdonald and
    Nicola Tonellotto},
  booktitle = {CIKM},
  pages = {3318--3322},
  title = {On Approximate Nearest Neighbour Selection for Multi-Stage Dense Retrieval},
  year = {2021}
}

@inproceedings{gerrish2011predicting,
  author = {Sean Gerrish and
    David M. Blei},
  booktitle = {ICML},
  pages = {489--496},
  title = {Predicting Legislative Roll Calls from Text},
  year = {2011}
}

@inproceedings{kraft2016embedding,
  author = {Peter Kraft and
    Hirsh Jain and
    Alexander M. Rush},
  booktitle = {EMNLP},
  pages = {2066--2070},
  title = {An Embedding Model for Predicting Roll-Call Votes},
  year = {2016}
}

@inproceedings{davoodi2020understanding,
  author = {Maryam Davoodi and
    Eric Waltenburg and
    Dan Goldwasser},
  booktitle = {ACL},
  pages = {5358--5368},
  title = {Understanding the Language of Political Agreement and Disagreement
in Legislative Texts},
  year = {2020}
}

@inproceedings{davoodi2022modeling,
  author = {Davoodi, Maryam  and
    Waltenburg, Eric  and
    Goldwasser, Dan},
  booktitle = {ACL},
  pages = {270--284},
  title = {{M}odeling {U.S.} State-Level Policies by Extracting Winners and Losers from Legislative Texts},
  year = {2022}
}

@inproceedings{mou2021align,
  author = {Xinyi Mou and
    Zhongyu Wei and
    Lei Chen and
    Shangyi Ning and
    Yancheng He and
    Changjian Jiang and
    Xuanjing Huang},
  booktitle = {ACL/IJCNLP},
  pages = {1236--1246},
  title = {Align Voting Behavior with Public Statements for Legislator Representation
Learning},
  year = {2021}
}

@inproceedings{liu2022generated,
  author = {Jiacheng Liu and
    Alisa Liu and
    Ximing Lu and
    Sean Welleck and
    Peter West and
    Ronan Le Bras and
    Yejin Choi and
    Hannaneh Hajishirzi},
  booktitle = {ACL},
  pages = {3154--3169},
  title = {Generated Knowledge Prompting for Commonsense Reasoning},
  year = {2022}
}

@inproceedings{sun2023recitationaugmented,
  author = {Zhiqing Sun and
    Xuezhi Wang and
    Yi Tay and
    Yiming Yang and
    Denny Zhou},
  booktitle = {ICLR},
  pages = {31171--31193},
  title = {Recitation-Augmented Language Models},
  year = {2023}
}

@inproceedings{wei2025instructrag,
  author = {Zhepei Wei and Wei-Lin Chen and Yu Meng},
  booktitle = {ICLR},
  pages = {66510--66533},
  title = {Instruct{RAG}: Instructing Retrieval-Augmented Generation via Self-Synthesized Rationales},
  year = {2025}
}

@inproceedings{lewis2019bart,
  author = {Lewis, Mike  and
    Liu, Yinhan  and
    Goyal, Naman  and
    Ghazvininejad, Marjan  and
    Mohamed, Abdelrahman  and
    Levy, Omer  and
    Stoyanov, Veselin  and
    Zettlemoyer, Luke},
  booktitle = {ACL},
  pages = {7871--7880},
  title = {{BART}: Denoising Sequence-to-Sequence Pre-training for Natural Language Generation, Translation, and Comprehension},
  year = {2020}
}

@inproceedings{reimers2019sentence,
  author = {Nils Reimers and
Iryna Gurevych},
  booktitle = {EMNLP-IJCNLP},
  pages = {3980--3990},
  title = {Sentence-BERT: Sentence Embeddings using Siamese BERT-Networks},
  year = {2019}
}

@inproceedings{lewis2020retrieval,
  author = {Patrick Lewis and
    Ethan Perez and
    Aleksandra Piktus and
    Fabio Petroni and
    Vladimir Karpukhin and
    Naman Goyal and
    Heinrich K{\""{u}}ttler and
    Mike Lewis and
    Wen{-}tau Yih and
    Tim Rockt{\""{a}}schel and
    Sebastian Riedel and
    Douwe Kiela},
  booktitle = {NeurIPS},
  pages={9459--9474},
  title = {Retrieval-Augmented Generation for Knowledge-Intensive {NLP} Tasks},
  year = {2020}
}

@inproceedings{baek2023knowledge,
  author = {Baek, Jinheon  and
Aji, Alham Fikri  and
Saffari, Amir},
  booktitle = {NLRSE},
  pages = {78--106},
  title = {Knowledge-Augmented Language Model Prompting for Zero-Shot Knowledge Graph Question Answering},
  year = {2023}
}

@inproceedings{wang2023can,
  author = {Wang, Heng and Feng, Shangbin and He, Tianxing and Tan, Zhaoxuan and Han, Xiaochuang and Tsvetkov, Yulia},
  booktitle = {NeurIPS},
  pages={30840--30861},
  title = {Can language models solve graph problems in natural language?},
  year = {2023}
}

@inproceedings{hu2022lora,
  author = {Edward J Hu and yelong shen and Phillip Wallis and Zeyuan Allen-Zhu and Yuanzhi Li and Shean Wang and Lu Wang and Weizhu Chen},
  booktitle = {ICLR},
  pages={12513--12525},
  title = {Lo{RA}: Low-Rank Adaptation of Large Language Models},
  year = {2022}
}

@inproceedings{feng2022political,
  author = {Feng, Shangbin  and
    Tan, Zhaoxuan  and
    Chen, Zilong  and
    Wang, Ningnan  and
    Yu, Peisheng  and
    Zheng, Qinghua  and
    Chang, Xiaojun  and
    Luo, Minnan},
  booktitle = {EMNLP},
  pages = {12022--12036},
  title = {{PAR}: Political Actor Representation Learning with Social Context and Expert Knowledge},
  year = {2022}
}

@inproceedings{hallucination,
    title = "A Comprehensive Survey of Hallucination in Large Language, Image, Video and Audio Foundation Models",
    author = "Sahoo, Pranab  and
      Meharia, Prabhash  and
      Ghosh, Akash  and
      Saha, Sriparna  and
      Jain, Vinija  and
      Chadha, Aman",
    booktitle = "EMNLP",
    year = "2024",
    pages = "11709--11724",
}

@article{guo2023gpt4graph,
  author = {Jiayan Guo and
Lun Du and
Hengyu Liu},
  journal = {CoRR},
  title = {GPT4Graph: Can Large Language Models Understand Graph Structured Data
? An Empirical Evaluation and Benchmarking},
  volume = {abs/2305.15066},
  year = {2023}
}

@article{clinton2004statistical,
  author = {Clinton, Joshua and Jackman, Simon and Rivers, Douglas},
  journal = {American Political Science Review},
  number = {2},
  pages = {355--370},
  title = {The statistical analysis of roll call data},
  volume = {98},
  year = {2004}
}

@article{feng2021kgap,
  author       = {Shangbin Feng and
                  Zilong Chen and
                  Qingyao Li and
                  Minnan Luo},
  title        = {KGAP: Knowledge Graph Augmented Political Perspective Detection in News Media},
  journal      = {CoRR},
  volume       = {abs/2108.03861},
  year         = {2021}
}

@article{grattafiori2024llama3herdmodels,
      title={The Llama 3 Herd of Models}, 
      author={Aaron Grattafiori and Abhimanyu Dubey and Abhinav Jauhri and Abhinav Pandey and Abhishek Kadian and Ahmad Al-Dahle and et al.},
      journal={CoRR},
      volume={abs/2407.21783},
      year={2024}
}

@article{Burnham_2025, 
      title={Stance detection: a practical guide to classifying political beliefs in text}, 
      volume={13}, 
      number={3}, 
      journal={Political Science Research and Methods}, 
      author={Burnham, Michael}, 
      year={2025}, 
      pages={611–628}
}

@article{gou2021knowledge,
  title={Knowledge distillation: A survey},
  author={Gou, Jianping and Yu, Baosheng and Maybank, Stephen J and Tao, Dacheng},
  journal={IJCV},
  volume={129},
  number={6},
  pages={1789--1819},
  year={2021},
  publisher={Springer}
}

@article{chung2024scaling,
  title={Scaling instruction-finetuned language models},
  author={Chung, Hyung Won and Hou, Le and Longpre, Shayne and Zoph, Barret and Tay, Yi and Fedus, William and Li, Yunxuan and Wang, Xuezhi and Dehghani, Mostafa and Brahma, Siddhartha and others},
  journal={JMLR},
  volume={25},
  number={70},
  pages={1--53},
  year={2024}
}

@article{aizawa2003information,
  title={An information-theoretic perspective of tf--idf measures},
  author={Aizawa, Akiko},
  journal={IPM},
  volume={39},
  number={1},
  pages={45--65},
  year={2003},
  publisher={Elsevier}
}

@article{robertson2009probabilistic,
  title={The probabilistic relevance framework: BM25 and beyond},
  author={Robertson, Stephen and Zaragoza, Hugo and others},
  journal={FTIR},
  volume={3},
  number={4},
  pages={333--389},
  year={2009},
  publisher={Now Publishers, Inc.}
}

@inproceedings{wang-han-2025-proprag,
    title = "{P}rop{RAG}: Guiding Retrieval with Beam Search over Proposition Paths",
    author = "Wang, Jingjin  and
      Han, Jiawei",
    booktitle = "EMNLP",
    year = "2025",
    pages = "6223--6238",
    ISBN = "979-8-89176-332-6"
}

@article{sowden2018quantifying,
  title        = {Quantifying compliance and acceptance through public and private social conformity},
  author       = {Sophie Sowden and Sofia Koletsi and Eva Lymberopoulos and Elisabeta Militaru and Caroline Catmur and Geoffrey Bird},
  journal      = {Consciousness and Cognition},
  volume       = {65},
  pages        = {359--367},
  year         = {2018},
}

@article{jiang2023mistral,
  title={Mistral 7B},
  author={Jiang, Albert Q. and Sablayrolles, Alexandre and Mensch, Arthur and Bamford, Chris and Devendra, Chaplot and Lample, Guillaume and Leach, Kevin and Stock, Pierre and Scao, Teven Le and others},
  journal={arXiv},
  year={2023}
}

@article{bi2024deepseek,
  title={DeepSeek LLM: Scaling Open-Source Language Models with Longtermism},
  author={Bi, Xiao and Chen, Deli and Chen, Guanting and Chen, Shanhuang and Dai, Damai and Deng, Chengqi and Ding, Hongyuan and Dong, Kai and E, Qiushi and others},
  journal={arXiv},
  year={2024}
}

@article{schafer2007collaborative,
  title={Collaborative Filtering Recommender Systems},
  author={Schafer, J. Ben and Frankowski, Dan and Herlocker, Jon and Sen, Shilad},
  journal={The Adaptive Web},
  pages={291--324},
  year={2007},
  publisher={Springer}
}

@inproceedings{he2020lightgcn,
  title={Lightgcn: Simplifying and powering graph convolution network for recommendation},
  author={He, Xiangnan and Deng, Kuan and Wang, Xiang and Li, Yan and Zhang, Yongdong and Wang, Meng},
  booktitle={SIGIR},
  pages={639--648},
  year={2020}
}

@inproceedings{zhang2024text,
  title={Text-like encoding of collaborative information in large language models for recommendation},
  author={Zhang, Yang and Bao, Keqin and Yan, Ming and Wang, Wenjie and Feng, Fuli and He, Xiangnan},
  booktitle={ACL},
  pages={9181--9191},
  year={2024}
}

@inproceedings{zhou2026diaglink,
  title={Diaglink: A dual-user diagnostic assistance system by synergizing experts with llms and knowledge graphs},
  author={Zhou, Zihan and Liu, Yinan and Xie, Yuyang and Wang, Bin and Yang, Xiaochun and Feng, Zezheng},
  booktitle={CHI},
  pages={1--28},
  year={2026}
}

\clearpage

\appendix

\section{EFFICIENCY STUDY}

We conduct experiments on RCVP to evaluate the running efficiency. To minimize the impact of platform differences, all experiments are performed using the GPT-3.5-Turbo API. As shown in Table~\ref{time}, the intermediate generation process is the primary factor contributing to the efficiency degradation. 
However, this process is inevitable when interacting with LLMs. 
Benefiting from the efficient utilization of the SLM, the intermediate generation time of PSL is shorter than that of existing generative approaches (\textit{e.g.}, RECITE, MindMap, and PEG). In future work, we plan to explore more efficient interaction strategies with LLMs.

\begin{table}[htbp]
\centering
{
\begin{tabular}{lll}
\toprule
\textbf{\textit{Method}} & 
\multicolumn{1}{c}{\textbf{\textit{RT}}} & 
\multicolumn{1}{c}{\textbf{\textit{Process}}} \\
\midrule
Vanilla               & 1.00  & FI \\
GKP                   & 2.10  & IG + FI \\
RECITE                & 3.11  & IG + FI \\
LangChain             & 1.67  & R + FI \\
InstructRAG           & 1.92  & R + IG + FI \\
KAPING                & 1.88  & R + FI \\
MindMap$_{route}$     & 1.95  & R + FI \\
MindMap$_{lang}$      & 4.99  & R + IG + FI \\
MindMap               & 13.9  & R + IG + FI \\
PEG$_{exp\_sum}$      & 2.52  & R + IG + FI \\
PEG$_{exp\_GTR}$      & 3.42  & R + IG + FI \\
PAA                   & 1.78  & FI \\
PSL                   & 2.48  & R + IG + FI \\
\bottomrule
\end{tabular}
}
\caption{Running efficiency analysis. Relative Time (RT) denotes the ratio of the model's running time using the GPT-3.5-Turbo interface to that of the baseline Vanilla GPT-3.5-Turbo. Process abbreviations: R denotes Retrieval, IG denotes Intermediate Generation, and FI denotes Final Inference.}
\label{time}
\end{table}

\section{EVALUATION}
For evaluation, we follow the setting of \cite{mou2024unifying}. Specifically, we provide multiple options in the prompt and instruct the model to output its choice. To handle cases where the options are not explicitly stated in the output, we employ regular expressions to match the answers effectively.

\begin{figure*}[t]
    \centering
    \begin{subfigure}[b]{0.32\textwidth}
        \centering
        \includegraphics[width=\textwidth]{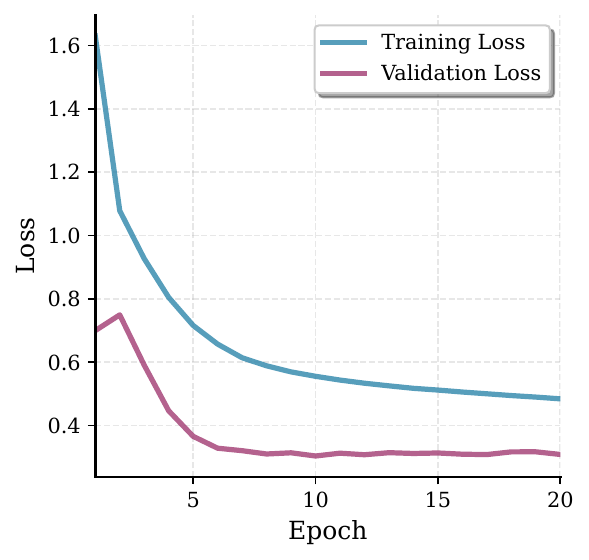}
        \caption{Loss curves}
        \label{fig:mlp_loss}
    \end{subfigure}
    \hfill
    \begin{subfigure}[b]{0.32\textwidth}
        \centering
        \includegraphics[width=\textwidth]{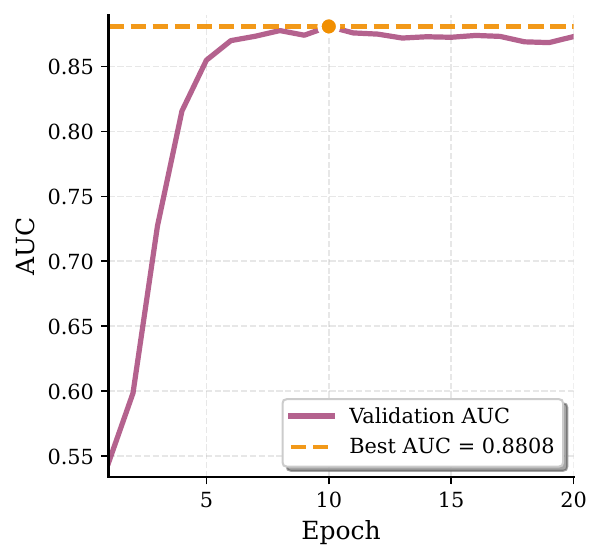}
        \caption{Validation AUC}
        \label{fig:mlp_auc}
    \end{subfigure}
    \hfill
    \begin{subfigure}[b]{0.32\textwidth}
        \centering
        \includegraphics[width=\textwidth]{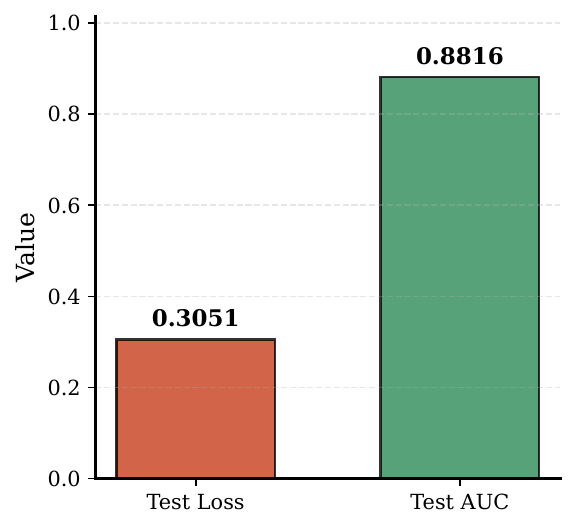}
        \caption{Test performance}
        \label{fig:mlp_test}
    \end{subfigure}
    \caption{MLP training curves and model performance over epochs.}
    \label{fig:mlp_training}
\end{figure*}

\begin{table}[t]
\centering
\begin{tabular}{ll}
\toprule
\textbf{Parameter} & \textbf{Value} \\
\midrule
Learning Rate & 5e-5 \\
Learning Rate Scheduler & Cosine \\
Epochs & 3.0 \\
Warmup Steps & 20 \\
Maximum Samples & 800 \\
Validation Split Ratio & 0.1 \\
Training Batch Size & 4 \\
Validation Batch Size & 1 \\
Gradient Accumulation Steps & 8 \\
Effective Batch Size & 32 (4×8) \\
Maximum Sequence Length & 2048 \\
Data Processing Threads & 16 \\
Precision Type & BF16 \\
\bottomrule
\end{tabular}
\caption{LoRA-Tuning parameters.}
\label{tab:lora_config}
\end{table}

\section{MLP TRAINING}

In this section, we detail the training procedure for the MLP. The training data is constructed from actor profiles. For each actor, every positive record embedding is paired with a randomly selected negative record embedding to form a training sample, \textit{i.e.}, (actor embedding, positive embedding, negative embedding). Aggregating these samples across all actors yields over 900{,}000 samples in total. From these, 500{,}000 samples are randomly selected to construct the dataset, using a random seed of 42, which is then split into training (70\%), validation (15\%), and test (15\%) subsets. The training parameters are summarized in Table~\ref{tab:mlp_config}. The model performance is evaluated by using AUC as the evaluation metric, while both loss and AUC are reported for the test set. Figure~\ref{fig:mlp_training} presents the training dynamics and final evaluation results.

\begin{table}[t]
\centering

\begin{tabular}{ll}
\toprule
\textbf{Parameter} & \textbf{Value} \\
\midrule
Batch Size & 40,960 \\
Max Epochs & 100 \\
Learning Rate & 0.0005 \\
Optimizer & Adam \\
Weight Decay & 0 \\
L2 Regularization & 1e-4 \\
Early Stopping Patience & 10 \\
Minimum Delta & 0 \\
\bottomrule
\end{tabular}
\caption{MLP training parameters.}
\label{tab:mlp_config}
\end{table}

\section{LoRA TUNING}

This section details the Low-Rank Adaptation (LoRA) fine-tuning method applied to LLMs. LoRA is a parameter-efficient fine-tuning technique that introduces trainable low-rank decomposition matrices alongside pre-trained weight matrices while keeping the original model parameters frozen. This approach significantly reduces the number of trainable parameters, lowering computational and storage costs while effectively adapting to downstream tasks. All experiments were conducted on 4 $\times$ NVIDIA A6000 GPUs. To conserve resources, the total number of training samples was limited to 800.
Table~\ref{tab:lora_config} summarizes the key parameter settings for LoRA fine-tuning.

\clearpage

\section{EXAMPLE OF STANCE}

\begin{figure}[htbp]
  \centering
  \promptbox{Sample stance of Seth Moulton}{
    1. **Previous Votes on Background Checks**: Seth Moulton has consistently voted 'yea' on the Enhanced Background Checks Act of 2019, indicating strong support for measures related to background checks for firearm sales. 2. **Legislative Focus**: Moulton's voting record shows a tendency to support legislation aimed at amending existing laws related to regulatory oversight, including labor relations and background checks. 3. **Consistency in Voting**: The repeated 'yea' votes on the same bill (Enhanced Background Checks Act of 2019) suggest a firm stance on the issue, which could carry over to his vote on H.B. 8 regarding background checks for every firearm sale. 4. **Context of Gun Control Legislation**: Moulton's history of supporting background check legislation may reflect a broader commitment to gun control measures, potentially influencing his decision on related bills. 5. **Political Implications**: As a member of Congress, Moulton may consider the political landscape and public opinion regarding gun control when deciding on votes, especially in light of his previous support for similar bills.
  }
  \label{fig2}
\end{figure}

\section{EXAMPLE OF POLITICAL BEHAVIOR PROFILE}

\begin{figure}[H]
  \centering
  \promptbox{Sample Profile of Lindsey Graham}{
    \{  \\
        \hspace{2em}"title": "Pechanga Band of Luiseno Mission Indians Water Rights Settlement Act", \\
        \hspace{2em}"description": "Pechanga Band of Luiseno Mission Indians Water Rights Settlement Act (Sec. 4) This bill authorizes, ratifies, and confirms the Pechanga Settlement Agreement, entered into by the Pechanga Band of Luiseno Mission Indians, the Rancho California Water District (RCWD), and the United States, except to the extent that the agreement is modified by or conflicts with this bill. (Sec. 5) The bill confirms water rights that must be held in trust by the United States on behalf of the tribe and its allottees. (Allottees are individuals who hold a beneficial real property interest in an Indian allotment that is located within the reservation and held in trust by the United States.) Allotted land is entitled to a just and equitable allocation of water from the water rights for irrigation and domestic purposes. Allottees may lease their land together with any water right. The tribe must enact a Pechanga Water Code that governs the storage, recovery, and use of the water rights, subject to the Department of the Interior's approval. Interior must administer the water rights until the water code is enacted and approved. (Sec. 7) The tribe and the United States (acting as trustee for the tribe and allottees) must waive all claims to water rights within the Santa Margarita River Watershed, except water rights recognized in the Pechanga Settlement Agreement and this bill. The tribe and the United States (acting as trustee for the tribe) waive specified claims against the RCWD. The tribe may waive claims against the United States regarding specified water rights and damages. The waivers in this bill are enforceable on the date Interior publishes specified findings regarding deposits, waivers, and approved agreements. (Sec. 8) Interior must provide the amounts necessary to fulfill the tribe's obligations under specified agreements regarding water infrastructure. (Sec. 9) The bill establishes the Pechanga Settlement Fund. The fund is to be used to carry out this bill. (Sec. 12) If Interior does not publish the findings required for enforcement of the waivers in this bill by April 30, 2021, or an alternative later date agreed to by the tribe and Interior, the provisions of this bill expire, related agreements are void, and funds are rescinded.", \\
        \hspace{2em}"summary": "The bill confirms that water rights shall be held in trust by the United States for the tribe and its allottees, who are individuals with beneficial interests in trust allotments within the reservation. The tribe must establish a Pechanga Water Code to regulate the storage, recovery, and use of these rights." \\
    \}, \\
    \{ \\
        \hspace{2em}"event\_text": "express intent to engage in diplomatic cooperation, including public policy support for Taiwan’s international participation, strengthening defense and security ties, and promoting economic exchanges", \\
        \hspace{2em}"target\_name": "taiwan"  \\
    \} 
  }
  \label{fig:your_label}
\end{figure}

\end{document}